\documentclass[letterpaper]{article} 
\usepackage{aaai25}  
\usepackage{times}  
\usepackage{helvet}  
\usepackage[hyphens]{url}  
\usepackage{graphicx} 
\usepackage{natbib}  
\usepackage{caption} 
\usepackage{algorithm}
\usepackage{algorithmic}

\usepackage{newfloat}
\usepackage{listings}
\DeclareCaptionStyle{ruled}{labelfont=normalfont,labelsep=colon,strut=off} 
\floatstyle{ruled}
\newfloat{listing}{tb}{lst}{}
\floatname{listing}{Listing}
\nocopyright 

\title{A Roadmap to Guide the Integration of LLMs in Hierarchical Planning}
\author {
    Israel Puerta-Merino,
    Carlos Núñez-Molina, 
    Pablo Mesejo,    
    Juan Fernández-Olivares
}
\affiliations{
    University of Granada, Spain\\
    Andalusian Institute of Data Science and Computational Intelligence (DaSCI)\\
    israelpm01@ugr.es, ccaarlos@ugr.es, pmesejo@go.ugr.es, faro@decsai.ugr.es
}

\usepackage{bibentry}

\usepackage{multirow}
\usepackage[table,xcdraw]{xcolor}

\begin{document}

\maketitle


\begin{abstract}

Recent advances in Large Language Models (LLMs) are fostering their integration into several reasoning-related fields, including Automated Planning (AP). However, their integration into Hierarchical Planning (HP), a subfield of AP that leverages hierarchical knowledge to enhance planning performance, remains largely unexplored. In this preliminary work, we propose a roadmap to address this gap and harness the potential of LLMs for HP. To this end, we present a taxonomy of integration methods, exploring how LLMs can be utilized within the HP life cycle. Additionally, we provide a benchmark with a standardized dataset for evaluating the performance of future LLM-based HP approaches, and present initial results for a state-of-the-art HP planner and \textit{LLM planner}. As expected, the latter exhibits limited performance (3\% correct plans, and none with a correct hierarchical decomposition) but serves as a valuable baseline for future approaches.

\end{abstract}

\section{Introduction}

Hierarchical Planning (HP) is a subfield within Automated Planning (AP) comprising planning methods that incorporate hierarchical knowledge. This hierarchical knowledge can be leveraged to speed up planning and, also, to integrate human expert problem-solving knowledge. While Large Language Models (LLMs) are being gradually integrated in various AI domains, including AP, their application to HP remains underexplored, with only a few studies tangentially addressing this topic \cite{yang2024oceanplan, dai2024optimal, tseimproving, song2023llm}. Our contribution within this work, therefore, is a roadmap to fill this gap, exploring the potential of LLMs for HP. 

We have analyzed existing literature on AP with LLMs, as well as current reviews \cite{pallagani2024prospects, huang2024understanding, valmeekam2022large}. Observing that most methods in AP that are similarly applicable to HP, we have classified these methods, building a taxonomy to bring HP closer to the existing LLM integration techniques. Our taxonomy classifies the integration methods according to two different dimensions: the \textit{Planning Process Role} (in which part of the HP life cycle is the LLM applied -- i.e. problem definition, plan elaboration or post-processing) and the \textit{LLM Improvement Strategy} (which LLM-based approach are used to improve the LLM performance -- i.e. giving extra knowledge or making multiple calls). Given that HP is a subset of AP, this classification is broadly applicable to AP as well.

To facilitate evaluation and comparisons among methods, we propose a standardized dataset and benchmarking framework based on the 2023 International Planning Competition (IPC-2023) HTN tracks, the most recent competition for HP solvers. Specifically, we suggest using the total-order track of the IPC-2023 dataset as a benchmark dataset.\footnote{Website of the IPC-2023: \url{https://ipc2023-htn.github.io}} As a baseline, we implement and evaluate a basic LLM Planner (direct planning using an LLM without any improvement strategies), the simplest method of our taxonomy, on this dataset. We use Llama-3.1-Nemotron-70B-Instruct \cite{wang2024helpsteer2preferencecomplementingratingspreferences}, one of the highest-performing LLMs available, and the model that we plan to use in subsequent experiments. We also provide the results from PandaDealer-agile-lama \cite{olz2023pandadealer}, the IPC-2023 Total-Order Satisficing track winner and state-of-the-art HP solver.

In summary, this work offers a roadmap to guide the future research in the integration of LLMs and HP, which is provided through two main contributions: the taxonomy of LLM in HP integration methods, illustrating the magnitude of this field and revealing how much work remains to be done; and the proposed benchmark, providing a tool for the evaluation and comparison of developed and future methods. We hope that this roadmap will inspire and guide future research in this promising yet underexplored field.

\section{Taxonomy}
\label{sec:classification}

To bridge the gap between HP and existing LLM Integration techniques, and to explore this expansive field, we propose a taxonomic framework that highlights the identified methods. This classification mainly draws from AP's state-of-the-art, so many of these methods also apply to AP. To provide context for the subsequent sections, we include illustrative examples of remarkable AP Integration methods from current literature. Each of the methods within this taxonomy represents a subset of techniques rather than a precise implementation, and may encompass multiple approaches. Furthermore, these methods are non-exclusive, meaning that planning agents can be designed to explore multiple combinations, making this a huge field to explore. This classification is intended to be a starting point, not a fixed framework, with room for exploration and refinement.

Our proposed taxonomy is structured along two dimensions: the \textit{Planning Process Role}, categorizing the stages of the HP life cycle where an LLM can be applied: problem definition, plan elaboration and post-processing (i.e., including plan translation and explanation to final user). And \textit{LLM Improvement Strategy}, which encompasses general strategies used to enhance LLM performance, which are applicable regardless of the role they are used for: knowledge enhancement and multiple calls. In this last dimension, despite several LLM Reasoning Strategies have been studied, we have focused our classification on the main different strategies that, we have observed, are commonly used in AP, as we consider that are the most interesting techniques to initially explore and evaluate in HP. These strategies mainly consist on using extra knowledge or augmenting the number of LLM executions.

\subsection{\textit{Planning Process Role}}

LLMs can assume various roles across the three general steps of a planning process: (1) problem definition, (2) plan elaboration, and (3) post-processing. Each step is further divided into the distinctive LLM integration methods observed in the literature. Table \ref{table:role} summarizes this classification.

\subsubsection{Problem Definition.} An HP Problem includes the same elements as an AP problem (\textit{Actions, Initial State} and \textit{Goal}), along with the high-level actions (\textit{Tasks}) that represents the hierarchical information about the environment. Each element can be generated using an LLM through two main approaches: \textit{Translation}, when all necessary information is explicitly and previously provided and we only want the LLM to restructure the provided information into a target format \cite{liu2023llm}; and \textit{Generation}, when the information is partially or implicitly provided, so the LLM is inferring or assuming missing information based on reasoning \cite{gestrin2024nl2plan}.

\subsubsection{Plan Elaboration}

We categorize this group based on the role of the LLM in the problem solving process:
\\
\begin{itemize}
    \item \textbf{LLM Planner}. In this basic setup, the LLM itself functions as the planner \cite{silver2022pddl}. While LLM improvement strategies can be applied, there is not external planner or explicit search process here, so the LLM the model is responsible for the entire planning process.
\\
    \item \textbf{Graph Search}. The LLM is embedded in the planner, which is done within an explicit search algorithm, where the LLM can perform one or more roles, like \textit{Node Expansion} (generating the next possible actions), \textit{Selection} (choosing the next action), \textit{Heuristic provision} (scoring states), \textit{Model Elicitation, Backtracking, Aggregation and Pruning}. Some remarkable architectures are RAP \cite{hao2023reasoning}, integrating an LLM into a MCTS Planner; GoT \cite{besta2024graph}, implementing an LLM in most of the mentioned roles to solve reasoning problems; and SayCan \cite{ahn2022can}, which utilizes an LLM as a probabilistic relevance (heuristic) scorer.

    \item \textbf{Planning Guidance}. Here, the LLM is external to the planner, but assists it by providing assistance. This guidance can be an \textit{initial plan} that the planner can refine \cite{valmeekam2023planning} or environment \textit{preferences} to narrow the search space \cite{sharan2023llm}.
\end{itemize}

\subsubsection{Post-Processing}

Once a plan is developed, an LLM can be used to refine it, by \textit{Translating the plan} into another data structure or language, typically an  executable format (e.g. for a robot to run it) or natural language \cite{liu2023llm}. Or \textit{Explaining the Plan}, where de LLM has to generate more detailed information, commonly in natural language, based on the provided plan \cite{simon2022tattletale}.

\begin{table}[t]
\centering
\begin{tabular}{llr}

\hline

\multirow{8}{*}{\textbf{Problem Definition}} & \multirow{4}{*}{Generation}         & \textit{Actions}              \\
                                             &                                     & \textit{Tasks}                \\
                                             &                                     & \textit{Initial State}        \\
                                             &                                     & \textit{Goal}                 \\ \cline{2-3} 
                                             & \multirow{4}{*}{Translation}        & \textit{Actions}              \\
                                             &                                     & \textit{Tasks}                \\
                                             &                                     & \textit{Initial State}        \\
                                             &                                     & \textit{Goal}                 \\ \hline

\multirow{10}{*}{\textbf{Plan Elaboration}}          
                                            & \multicolumn{2}{c}{\multirow{2}{*}{\textit{LLM Planner}}}                  \\\\ \cline{2-3} 
                                             & \multirow{7}{*}{Graph Search}        & \textit{Expansion}    \\
                                             &                                     & \textit{Selection}     \\
                                             &                                     & \textit{Elicitation} \\
                                             &                                     & \textit{Backtracking}         \\
                                             &                                     & \textit{Heuristic}            \\
                                             &                                     & \textit{Aggregation}          \\
                                             &                                     & \textit{Pruning}             \\ \cline{2-3} 
                                             & \multirow{2}{*}{Guidance}  & \textit{Preferences} \\
                                             &                                     & \textit{Initialization}       \\ 
                                             \hline
\multirow{2}{*}{\textbf{Post-Processing}}     & \multicolumn{2}{c}{\textit{Translating the Plan}}                            \\
                                             & \multicolumn{2}{c}{\textit{Explaining the Plan}}                            \\ \hline
\end{tabular}
\caption{Summary of the possible roles that an LLM can perform during the HP life cycle. This classification is detailed in the \textit{Planning Process Role} section.}
\label{table:role}
\end{table}

\subsection{\textit{LLM Improvement Strategies}}

When performing any task using an LLM, it can be directly executed to obtain an output. However, their results are often suboptimal. To address this, there are some strategies to enhance the LLM performance which have been used in several reasoning-related fields, as math, question answering or AP. Building on this foundation, we have identified the strategies commonly used in AP, and extended them to the specific case of HP. In this section, we propose a classification of strategies that can be employed in HP to improve the LLM performance, regardless of the role they are used for.




LLM performance can be enhanced by either providing more knowledge about the problem or increasing the number of LLM calls used during problem-solving. Each of these approaches is also categorized into the distinctive strategies. This classification is detailed along next subsections and summarized in Table \ref{table:technique}. Note that these strategies are neither mandatory nor mutually exclusive, meaning that it is possible to use none, some, or all of them simultaneously.


\subsubsection{Knowledge Enhancement}

We can split the LLM knowledge enhancement strategies into two main perspectives: with previous knowledge (providing additional information relevant to the problem before starting to solve it) or through feedback (the model is iteratively provided with extra information based on its previous outputs, while solving the task). Both perspectives, moreover, may have different approaches depending on the location where the knowledge is applied: 

\begin{itemize}
    \item \textbf{Previous}. This approach involves providing extra information to the LLM before its execution. \textit{Fine-tuning} is the traditional method in the field of deep learning, which involves adjusting the model's internal weights with supplementary data \cite{Pallagani_Muppasani_Murugesan_Rossi_Horesh_Srivastava_Fabiano_Loreggia_2022}.  Alternatively, knowledge can be directly provided through the model’s prompt, which is a more flexible and accessible option than fine-tuning, as it does not require a training process. Providing extra information is achieved through input-output examples (shots) that enhance generated responses \cite{song2023llm}. If the shots provide useful environmental information (e.g., domain information) this is called \textit{in-context prompting}; if they only aim to elicit an specific output structure, it is referred to as \textit{out-of-context prompting}. Furthermore, we can also provide knowledge about how to reason, about the problem itself. \textit{Chain of Thoughts} (CoT) \cite{Wei_Wang_Schuurmans_Bosma_Ichter_Xia_Chi_Le_Zhou_2023} is the best example of this, a different prompting strategy that encourages the LLM to generate a reasoning process before providing the final answer. This is typically achieved by using \textit{few-shot prompting} with reasoning examples, though reasoning can also be elicited without examples, known as \textit{Zero-shot CoT} \cite{kojima2022large}.

     \item \textbf{Feedback.} Unlike the previous paradigm, here the extra information is provided during the task solving process: rather than simply accepting the output of an LLM execution, we can iteratively improve the result of the LLM by providing it feedback based on its previous outputs. This feedback can come from various sources (humans, the environment, an external module, another LLM, etc.).  \textit{Reinforcement Learning} is the classical method of learning from feedback in machine learning, where the knowledge is applied by modifying the model’s internal weights, based on a reward score \cite{yao2020keep}. Alternatively, feedback can be used to dynamically adjust and improve the prompt (\textit{Prompt Correction}), where prominent examples in the literature are Self-Refine \cite{madaan2024self} and Reflexion \cite{Shinn_Cassano_Berman_Gopinath_Narasimhan_Yao_2023}. Finally, the model's learning can also be represented through an external \textit{Memory Module}, being Voyager a notable architecture that uses this approach \cite{Wang_Xie_Jiang_Mandlekar_Xiao_Zhu_Fan_Anandkumar_2023}.
\end{itemize}
    
    
    \subsubsection{Multiple Calls} 
    To improve the LLM's performance through multiple calls, there are two main perspectives: the problem can be divided into simpler sub-problems that the LLM solve on each call (Decomposition); or the LLM solves the whole problem several times and we take advantage of the varied information generated by the multiple outputs (Revision). Both perspectives can also be addressed through two different approaches, depending on whether an output is used for the next call or not (\textit{Sequential} or \textit{Parallel} calls).

\begin{itemize}
    \item \textbf{Decomposition.} A problem can be sequentially decomposed: each LLM call's output generates an intermediate step that is utilized for the next call's input. For instance, to generate a plan, we could iteratively ask the LLM to only generate the next action of a partial plan, than attempting to generate the entire plan in a single call \cite{Huang_Abbeel_Pathak_Mordatch_2022}. Otherwise, the problem can be decomposed in parallel: each LLM call can returns a list simpler sub-problems which can be independently solved (i.e. the classical Divide and Conquer strategy). This is illustratively used in Generative Agents \cite{park2023generative}.
    
    \item \textbf{Revision.} We can make the multiple calls to sequentially refine the LLM's output: we ask the LLM to initially give us a complete but simple (general) solution that iteratively becomes more detailed through the calls \cite{liu2024delta}. Otherwise, we can do a parallel approach by asking the LLM to entirely solve the problem multiple times and, then, combine the different results into a final output. Self-consistency \cite{wang2022self} is a prominent approach, which combines the several outputs by selecting the most common response, but various criteria can be used to choose the final output.
\end{itemize}

\begin{table}[t]
\centering
\begin{tabular}{ccr}
\hline
\multirow{6}{*}{\textbf{Knowledge}}   & \multirow{3}{*}{Previous}      & \textit{Fine-tuning}            \\
                                      &                                & \textit{Prompting}              \\
                                      &                                & \textit{Chain of Thoughts}      \\ \cline{2-3} 
                                      & \multirow{3}{*}{Feedback}      & \textit{Reinforcement} \\
                                      &                                & \textit{Prompt Correction}      \\
                                      &                                & \textit{Memory}                 \\ \hline
\multirow{4}{*}{\textbf{Multi-Calls}} & \multirow{2}{*}{Decomposition} & \textit{Sequential Calls}       \\
                                      &                                & \textit{Parallel Calls}         \\ \cline{2-3} 
                                      & \multirow{2}{*}{Revision}      & \textit{Sequential Calls}       \\
                                      &                                & \textit{Parallel Calls}         \\ \hline
\end{tabular}
\caption{Summary of the possible strategies that can be followed to improve the LLM's performance within the HP life cycle, regardless of the specific role. This classification is detailed in the \textit{LLM Improvement Strategies} section.}
\label{table:technique}
\end{table}

\section{Benchmark}

Hierarchical Task Networks (HTN) Planning \cite{nau2003shop2} is the most widely used and studied approach within the HP field, making it a suitable reference point for experimenting with and evaluating various LLM integration methods. We propose using the total-order track from IPC-2023, which contains 22 domains, each with dozens of problems. A detailed breakdown of this information, along with a summary of the experimental results, is provided in Table \ref{table:experimentation}.

To assess and compare the performance of future implementations, we establish two reference points. First, we consider the winner of the IPC-2023 Total-Order Satisficing track, PandaDealer-agile-lama \cite{olz2023pandadealer} as an upper bound, as it represents the current state of the art in HP solvers. Second, we include as baseline a basic LLM Planner (i.e. using an LLM directly to plan without providing any LLM Improvement Strategy) representing a lower bound for LLM Integration. This approach, being the simplest method in our taxonomy, serves as a foundational point of comparison and a starting point for the roadmap.

\subsection{\textit{Execution and Experimentation Considerations}}

The PandaDealer score within the dataset have been sourced from the published IPC-2023 results. This score represents the ratio $C^*/C$ between the cost of a reference plan ($C^*$) and the best obtained plan ($C$). Due to the unavailability of these reference plans, we could not compute the same score for the LLM Planner. Nonetheless, this score serves as a reliable reference for assessing the quality of PandaDealer's performance. For generative plans, however, additional properties need to be evaluated, such as semantic coherence within the plan (which does not need to be measured in symbolic planners). To address this, we adopted alternative metrics more suitable for analyzing the quality of generative plans: \textit{Plan Feasibility} (the plan is syntactically correct),  \textit{Plan Correctness} (it is executable and reaches a goal state), \textit{Decomposition Feasibility} (the hierarchical decomposition of the plan is syntactically correct) and \textit{Decomposition Correctness} (the decomposition aligns with the generated plan).

We utilize Llama-3.1-Nemotron-70B-Instruct, one of the highest-performing LLMs available \cite{wang2024helpsteer2preferencecomplementingratingspreferences}, as the LLM Planner. We utilize the official IPC verifier to score the generated plans along the different proposed metrics. The execution and validation source codes, as well as the generated plans, and obtained results are allocated on GitHub. URL: \url{https://github.com/Corkiray/HTN-LLM}.

\subsection{\textit{Results Discussions}}

In this preliminary work, we have obtained the results for 15 out of the 23 domains in the dataset, not the complete set, due to temporal and computational limitations. Nevertheless, these results are sufficient to illustrate the notably low performance of the LLM Planner, which is expected given the simplicity of the method and the absence of any LLM Improvement Strategies. As shown in Table \ref{table:experimentation}, the LLM Planner fails to generate feasible plans in nearly 70\% of the problems, highlighting the LLM's limited ability to interpret and adhere to a specific format. Even more remarkable is the proportion of correct plans: with only 4\% (i.e. 13\% of the feasible plans), it shows the difficulty that a basic LLM encounters in planning correctly. A similar trend is observed in the number of feasible decompositions which, with only 3\% correct, indicates a significant performance drop in handling a specific format with increasing complexity. Interestingly, correct plans and feasible decompositions are often disjoint (i.e. the LLM achieves either one result or the other, but not both). Consequentially, the LLM has been unable to produce a correct hierarchical decomposition for any problem. This observation is explainable considering the LLM's inherent architecture as a transformer, which operates within a static computational capacity. An LLM cannot dynamically adjust its processing time based on problem complexity, and as a result, its ability to address multiple demanding requirements simultaneously (i.e. plan correctness and decomposition feasibility) is limited.

\begin{table}[t]
\centering
\begin{tabular}{lcrrrrr}
\multicolumn{1}{c}{}          &            & \textbf{Panda} & \multicolumn{4}{c}{\textbf{LLM Planner}} \\
\multicolumn{1}{c}{\textbf{Domain}}    & N   & Score                & FP       & CP       & FD       & CD      \\ \hline
\textit{Assembly}             &     30       &      0.89        &        15 &	3&	0&	0      \\
\textit{Barman}               &    20        &       0.78               &       4	&1	&0	&0       \\
\textit{Blocks}          &    30        &        0.77              &      9	&0	&1	&0     \\
\textit{Depots}               &      30      &         0.90             &     4	&0	&0	&0         \\
\textit{Factories}            &       20    &         0.67             &        6	&0	&0	&0      \\
\textit{Freecell}             &       60     &          0.13            &       38	&18	&0	&0        \\
\textit{Hiking}               &      30      &        0.83              &      0	&0	&0	&0         \\
\textit{lamps}                &      30      &         0.48             &       5	&1	&1	&0       \\
\textit{Logistics}            &      80      &      0.98                &       14	&1&	1	&0    \\
\textit{Multiarm}             &     74       &         0.95             &     30	&0	&2	&0       \\
\textit{Robot}                &     20       &         0.92             &      13	&1	&5	&0      \\
\textit{Satellite}            &     20       &        0.92              &         11	&0	&2	&0      \\
\textit{Towers}               &    20        &         0.65             &        5&	0&	0&	0        \\
\textit{Transport}            &     40       &        0.73              &      10&	0&	4&	0         \\
\textit{Woodwork}          &     30       &        0.69              &       26	&0	&0	&0       \\
\hline
\multicolumn{1}{l}{\textit{\textbf{Total}}}                &      614      &        14.42             &     190     &    25      &    16      &    0     \\
\hline 
\multicolumn{2}{c}{\textbf{Average:}} &        \textbf{0.80}	  &     \textbf{0.31}     &     \textbf{0.04}     &    \textbf{0.03}      &   \textbf{0}   \\ \hline  
\end{tabular}
\caption{Summary of the information provided by the benchmark, listing the number of problems in each domain of the dataset and the performance reported by each planner. The score of PandaDealer is the one provided by the IPC-2023 results. The scores utilized for the LLM Planner are explained in the \textit{Benchmark} section, representing the number of feasible plans (FP), correct plans (CP), feasible decompositions (FD) and correct decompositions (CD).}
\label{table:experimentation}
\end{table}

\section{Conclusion}

In this work, we propose a roadmap to guide future research in the integration of LLMs and HP, as this remains a largely unexplored field. This roadmap is centered on two key contributions: a taxonomy and a benchmark. The taxonomy categorizes the main integration methods, structured along two dimensions: the first one considers the various roles that an LLM could fulfill within the HP life cycle. The second one focuses on strategies to enhance LLM performance, irrespective of the specific role in which they are applied. The benchmark introduces a dataset and provides initial results that serves as reference in subsequent experimentation. These results include the performance of a state-of-the-art HP solver and an LLM Planner using one of the best-performing LLMs available, but without leveraging any improvement strategy. As expected, the results reveal the LLM's limited performance in solving HP problems, but they establishes a baseline to evaluate succeeding improvements. Promising future directions include exploring the planning capabilities of an LLM Planner augmented with improvement Strategies to overcome the limitations identified in this study. Furthermore, exploring integration in additional aspects of the HP life cycle, such as Plan Monitoring or Exception Management, is beneficial to expand the boundaries of the proposed taxonomy. Finally, developing new architectures within the outlined Planning Process Roles offers a pathway to systematically investigate and advance this unexplored field.

\section{Acknowledgment}

This work has been partially funded by the Grant PID2022-142976OB-I00, funded by MICIU/AEI/ 10.13039/501100011033 and by “ERDF/EU”.

\bibliography{aaai25}

\end{document}